%% file: ijcai19.tex
\documentclass{article}
\usepackage{ijcai19}

\usepackage{times}
\usepackage{soul}
\usepackage[hidelinks]{hyperref}
\usepackage{url}
\usepackage[utf8]{inputenc}
\usepackage[small]{caption}
\usepackage{graphicx}
\usepackage{amsmath}
\usepackage{booktabs}
\usepackage{algorithm}
\usepackage[htt]{hyphenat}
\usepackage[symbol]{footmisc}

\usepackage{algorithmic}
\newcommand{\minenet}{MineRL}

\usepackage[textsize=scriptsize]{todonotes}
\title{\minenet: A Large-Scale Dataset of Minecraft Demonstrations}
 
\author{
William H. Guss\footnote{Equal contribution.}\footnote{Contact Author.}\and
Brandon Houghton\footnotemark[1]\and
Nicholay Topin\and
Phillip Wang\and 
Cayden Codel\and
Manuela Veloso\And
Ruslan Salakhutdinov\\
\affiliations
Carnegie Mellon University, Pittsburgh, PA 15289, USA\\
\emails
\{wguss, bhoughton, ntopin, pkwang, ccodel, mmv, rsalakhu\}@cs.cmu.edu
}

\begin{document}

\maketitle

\begin{abstract}
    The sample inefficiency of standard deep reinforcement learning methods precludes their application to many real-world problems.
        Methods which leverage human demonstrations require fewer samples but have been researched less.
    As demonstrated in the computer vision and natural language processing communities, large-scale datasets have the capacity to facilitate research by serving as an experimental and benchmarking platform for new methods. 
        However, existing datasets compatible with reinforcement learning simulators do not have sufficient scale, structure, and quality to enable the further development and evaluation of methods focused on using human examples. 
    Therefore, we introduce a comprehensive, large-scale, simulator-paired dataset of human demonstrations: MineRL. 
        The dataset consists of over 60 million automatically annotated state-action pairs across a variety of related tasks in Minecraft, a dynamic, 3D, open-world environment. 
    We present a novel data collection scheme which allows for the ongoing introduction of new tasks and the gathering of complete state information suitable for a variety of methods. 
        We demonstrate the hierarchality, diversity, and scale of the MineRL dataset. 
        Further, we show the difficulty of the Minecraft domain along with the potential of MineRL in developing techniques to solve key research challenges within it. 
\end{abstract}

\input{sections/introduction}

\input{sections/background}

\input{sections/methodology}
\input{sections/results}

\input{sections/related_works}
\input{sections/conclusion}

{
\small
\bibliographystyle{named}
\bibliography{ijcai19}
}

\end{document}

%% file: sections/introduction.tex

\section{Introduction}

\begin{figure}
    \begin{center}
        \vspace{-1pt}
        \includegraphics[width=0.47\textwidth]{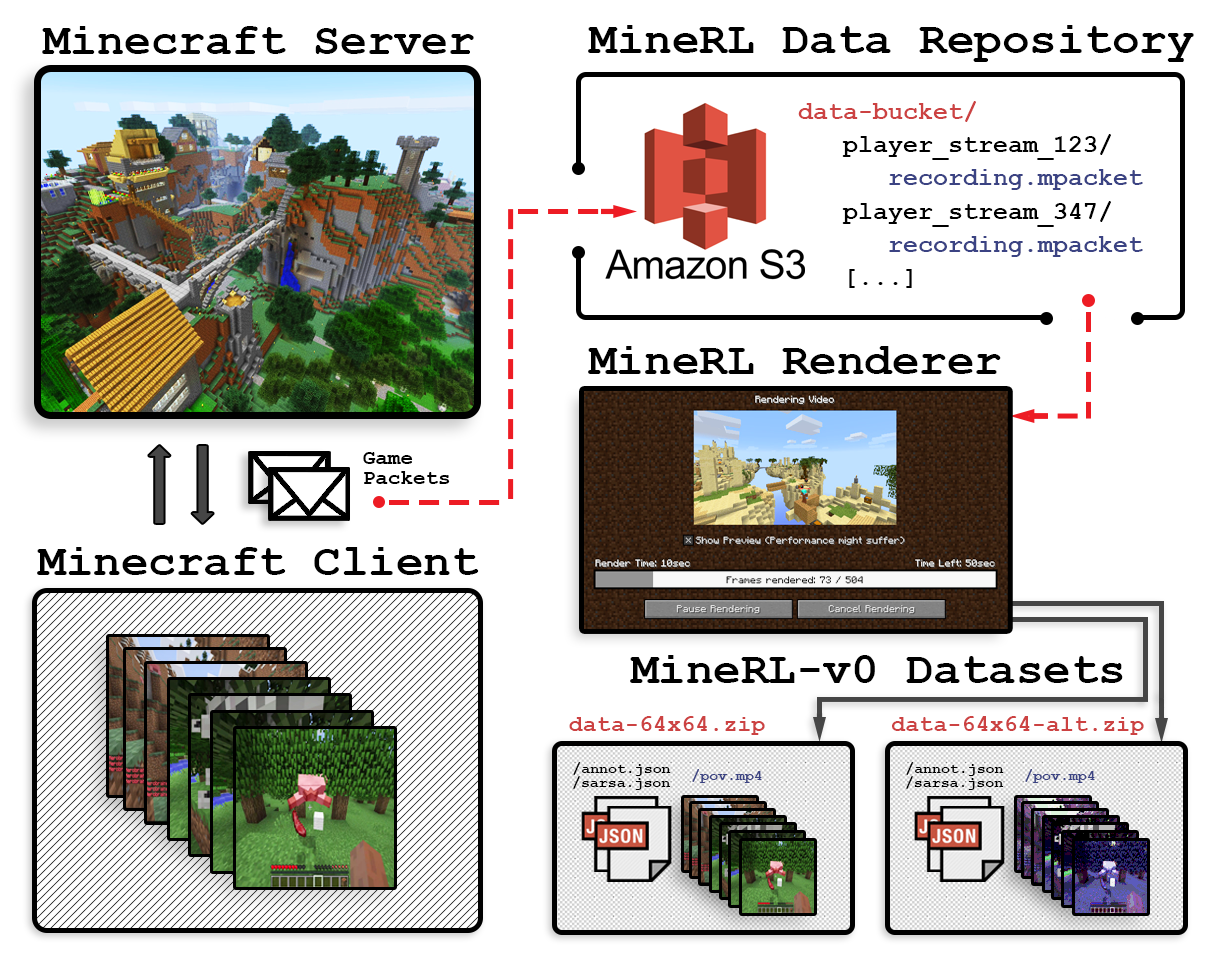}
        \vspace{-3pt}
        \caption{A diagram of the \minenet{} data collection platform. Our system renders demonstrations from packet-level data, so the game state and rendering parameters can be changed.} \label{fig:platform_diagram}
    \end{center}
    \vspace{-22pt}
\end{figure}

    As deep reinforcement learning (DRL) methods are applied to increasingly difficult problems, the number of samples used for training increases.
    For example, Atari 2600 games~\cite{bellemare2013arcade} have been used to evaluate DQN~\cite{mnih2015human}, A3C~\cite{mnih2016asynchronous}, and Rainbow DQN, which each require from 44 to over 200 million frames (200 to over 900 hours) to achieve human-level performance~\cite{hessel2018rainbow}.
    On more complex domains: OpenAI Five utilizes 11,000+ years of Dota 2 gameplay~\cite{openai_2018}, AlphaGoZero uses 4.9 million games of self-play in Go~\cite{silver2017mastering}, and AlphaStar uses 200 years of Starcraft~II gameplay~\cite{deepmind}.
    
    This inherent sample inefficiency precludes the application of standard DRL methods to real-world problems without leveraging 
    data augmentation techniques~\cite{tobin2017domain}, \cite{andrychowicz2018learning}, 
    domain alignment methods~\cite{wang2018pix2pixHD}, or %
    carefully designing real-world environments to allow for the required 
    number of trials~\cite{levine2018learning}.
    Recently, techniques leveraging trajectory examples, such as imitation learning and Bayesian reinforcement learning methods, have been successfully applied to older benchmarks and real-world problems where samples from the environment are costly.
    
    However, these techniques are still not sufficiently sample efficient for a large class of complex real-world domains.

    As noted %
    by \cite{kurin2017atari}, several subfields of machine learning have been catalyzed by the introduction of datasets and efficient large-scale data collection schemes, 
    such as Switchboard~\cite{godfrey1992switchboard} and ImageNet~\cite{deng2009imagenet}.
    Though the reinforcement learning community has created an extensive range of benchmark simulators,
    there is currently a lack of large-scale labeled datasets of human demonstrations for domains with a broad range of structural constraints and tasks.
          
    \emph{Therefore, we introduce \minenet{}, a large-scale, dataset of {over 60 million state-action pairs} of human demonstrations across 
    a range of related tasks in Minecraft}.
    To capture the diversity of gameplay and player interactions in Minecraft, \minenet{} includes six tasks with a variety of research challenges including open-world multi-agent interactions, 
            long-term planning,
            vision,
            control,
            and navigation,
            as well as explicit and implicit subtask hierarchies.
    We provide implementations of these tasks as sequential decision-making environments in an existing Minecraft simulator. 
    \emph{Additionally, we introduce a novel platform and methodology for the continued collection of human demonstrations in Minecraft.}
        As users play on our publicly available game server, we record packet-level information, which allows perfect reconstruction of each player's view and actions. This platform enables the addition of new tasks to the \minenet{} dataset and automatic annotation to complement current and future methods applied to Minecraft.
        Demo videos and more details about the dataset can be found at \url{http://minerl.io}.

%% file: sections/background.tex

\section{Environment: Minecraft}

\begin{figure}
    \begin{center}
        \includegraphics[width=0.47\textwidth]{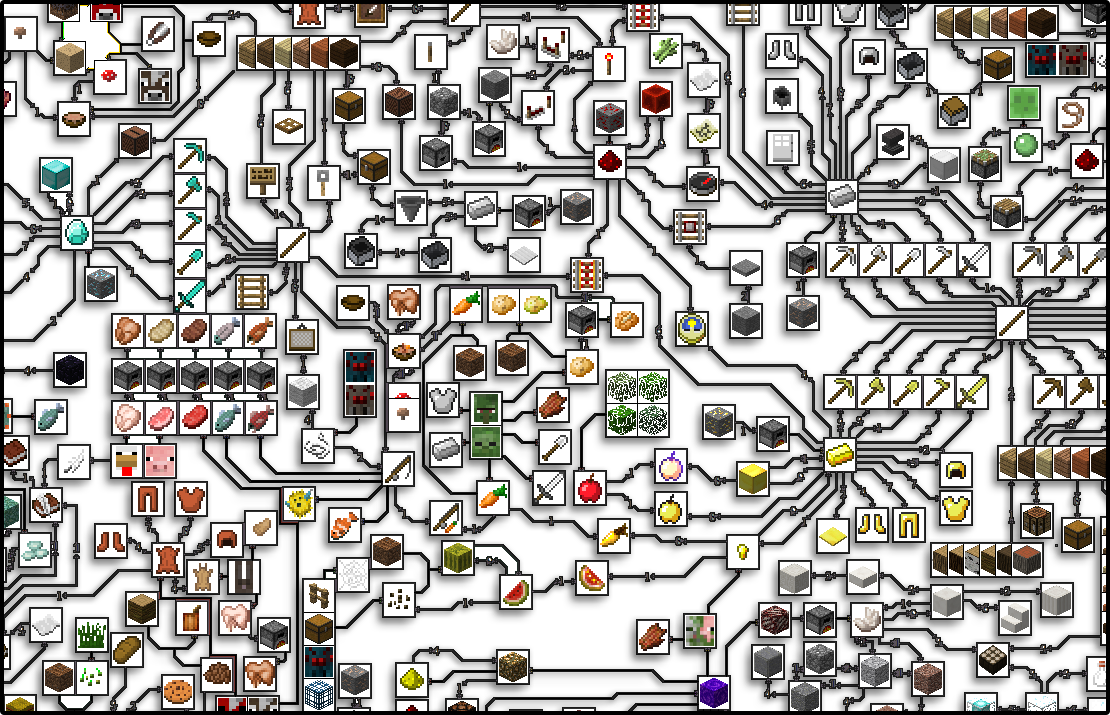}
        \vspace{2pt}
        \caption{A subset of the Minecraft item hierarchy (totaling 371 unique items). Each node is a unique Minecraft item, block, or non-player character, and a directed edge between two nodes denotes that one is a prerequisite for another. Each item presents is own unique set of challenges, so coverage of the full hierarchy by one player takes several hundred hours.} \label{fig:hierarchicality}
    \end{center}
    \vspace{-19pt}
\end{figure}

\subsection{Description} 
Minecraft is a compelling domain for the development of reinforcement and imitation learning based methods because of the unique challenges it presents:  
Minecraft is a 3D, first-person, open-world game centered around the gathering of resources and creation of structures and items. It can be played in a single-player mode or a multi-player mode, where all players exist in and affect the same world. Games are played across many sessions, for tens of hours total, per player. Notably, the procedurally generated world is composed of discrete blocks which allow modification; over the course of gameplay, players change their surroundings by gathering resources (such as wood from trees) and constructing structures (such as shelter and storage).

As an open-world game, Minecraft has no single definable objective. Instead, players develop their own subgoals which form a multitude of natural hierarchies. Though these hierarchies can be exploited, their size and complexity contribute to Minecraft's inherent difficulty. One such hierarchy is that of \emph{item collection}: for a large number of objectives in Minecraft, players must create specific tools, materials, and items which require the collection of a strict set of requisite items. The aggregate of these dependencies forms a large-scale task hierarchy (see Figure \ref{fig:hierarchicality}).

In addition to obtaining items, implicit hierarchies emerge through other aspects of gameplay. For example, players (1) construct structures to provide safety for themselves and their stored resources from naturally occurring enemies and (2) explore the world in search of natural resources, often engaging in combat with non-player characters. Both of these gameplay elements have long time horizons and exhibit flexible hierarchies due to situation dependent requirements (such as farming a certain resource necessary to survive, enabling exploration to then gather another resource, and so on).

\subsection{Existing Interest} With the development of Malmo~\cite{johnson2016malmo}, a simulator for Minecraft, the environment has garnered great research interest:
    \cite{shu2017hierarchical}, \cite{tessler2017deep}, and \cite{oh2016control} have leveraged Minecraft's massive hierarchality and expressive power as a simulator to make great strides in 
        language-grounded, interpretable multi-task option-extraction, hierarchical lifelong learning, and active perception. 
    However, much of the existing research utilizes toy tasks 
        in Minecraft, often restricted to 2D movement, discrete positions, or artificially confined maps unrepresentative of the intrinsic complexity that human players typically face. %
    These restrictions reflect the difficulty
        of the domain as well as the inability of current approaches to cope with fully embodied human state- and action-spaces and the complexity exhibited in optimal human policies.
    This inability is further evidenced by the large-body of work developed on Minecraft-like domains which specifically captures restricted subsets of the features of Minecraft 
    \cite{salge2014changing}, \cite{andreas2017modular}, \cite{liu2017interactive}.

Bridging the gap between these restricted Minecraft environments and the full domain encountered by humans is a driving force behind the development of \minenet{}.
    To do this, \minenet{}-v0 captures core aspects of 
        Minecraft that have motivated its use as a research domain, including its hierarchality and its large family of intrinsic subtasks. 
    At the same time, \minenet{}-v0 provides the human priors and rich, automatically generated meta-data necessary to enable current and future research to tackle the full Minecraft domain.

%% file: sections/methodology.tex

\vspace{-7pt}
\section{Methods: \minenet~Data Collection Plaform}
Classification and natural language datasets have benefited greatly from the existence of data collection platforms like Mechanical Turk, but, in contrast, the collection of gameplay data usually requires the implementation of a new platform and user acquisition scheme for each game. To that end, we introduce the first end-to-end platform for the collection of player trajectories in Minecraft, enabling the construction of the \minenet-v0 dataset. As shown in Figure~\ref{fig:platform_diagram}, our platform consists of 
    (1) \emph{a public game server and website}, where we obtain permission to record trajectories of Minecraft players in natural gameplay;
    (2) \emph{a custom Minecraft client plugin}, which records all packet level communication between the client and the server, so we can re-simulate and re-render human demonstrations with modifications to the game state and graphics; 
    and (3) \emph{a data processing pipeline}, which enables us to produce automatically annotated datasets of task demonstrations.

    
    \paragraph{Data Acquisition.}
   Minecraft players find the \minenet~server on standard Minecraft server lists. 
        Players first use our webpage to provide IRB consent for having their gameplay anonymously recorded. They then download a plugin for their Minecraft client which records and streams users' client-server game packets to the \minenet~data repository.
    When playing on our server, users select a stand-alone task to complete and receive in-game currency proportional to the amount of reward obtained. 
        For the \texttt{Survival} game mode (where there is no known reward function), players receive rewards only for duration of gameplay so as not to impose an artificial reward function.
        We implement each of these stand-alone tasks in Malmo.


    \paragraph{Data Pipeline.}
    Our data pipeline enables the continued expansion of the structured information accompanying \minenet{} dataset releases; it allows us to resimulate, modify, and augment recorded trajectories into several algorithmically consumable formats. 
        The pipeline serves as an extension to the core Minecraft game code and synchronously resends 
            each recorded packet from the \minenet{} data repository to a Minecraft client using our custom API for automatic annotation and game-state modification. 
        This API allows us to add annotations based on any aspect of the game state accessible from existing Minecraft simulators. 

    \paragraph{Extensibility.}
    Our aim is to use our platform to provide an exhaustive and broad set of multi-task datasets (beyond \minenet-v0) paired with RL environments, spanning natural language, embodied reasoning, hierarchical planning, and multi-agent cooperation.
        The modular design of the server allows us to obtain data for a growing number of stand-alone tasks. 
        Furthermore, the in-game economy and server community create consistent engagement from the user-base allowing us to collect data at a growing rate without incurring additional costs.
    The modularity, simulator compatibility, and configurability of the data pipeline also allows new datasets to be created to compliment new techniques leveraging human demonstrations.
        For example, it is possible to conduct large-scale generalization studies by repeatedly re-rendering the data with different constraints: altered lighting, camera positions (embodied and non-embodied), and other video rendering conditions; the injection of artificial noise in observations, rewards, and actions; and game hierarchy rearrangment (swapping the function and semantics of game items).

%% file: sections/results.tex

\section{Results: \minenet-v0}

In this section, we introduce and analyze the \minenet-v0 dataset. We first give details about the dataset including its size, form, and packaging. Then we indicate the wide applicability of this initial release by giving a detailed account of the included tasks families, followed by an analysis of the data quality, coverage, and hierarchicality. To frame the usefulness of the \minenet-v0 dataset, in Section~\ref{sec:difficulty}, we demonstrate the difficulty of our tasks with respect to out-of-the-box methods and the performance increase achieved through basic imitation learning techniques using \minenet-v0.

\begin{figure}
    \begin{center}
        \includegraphics[width=0.49\textwidth]{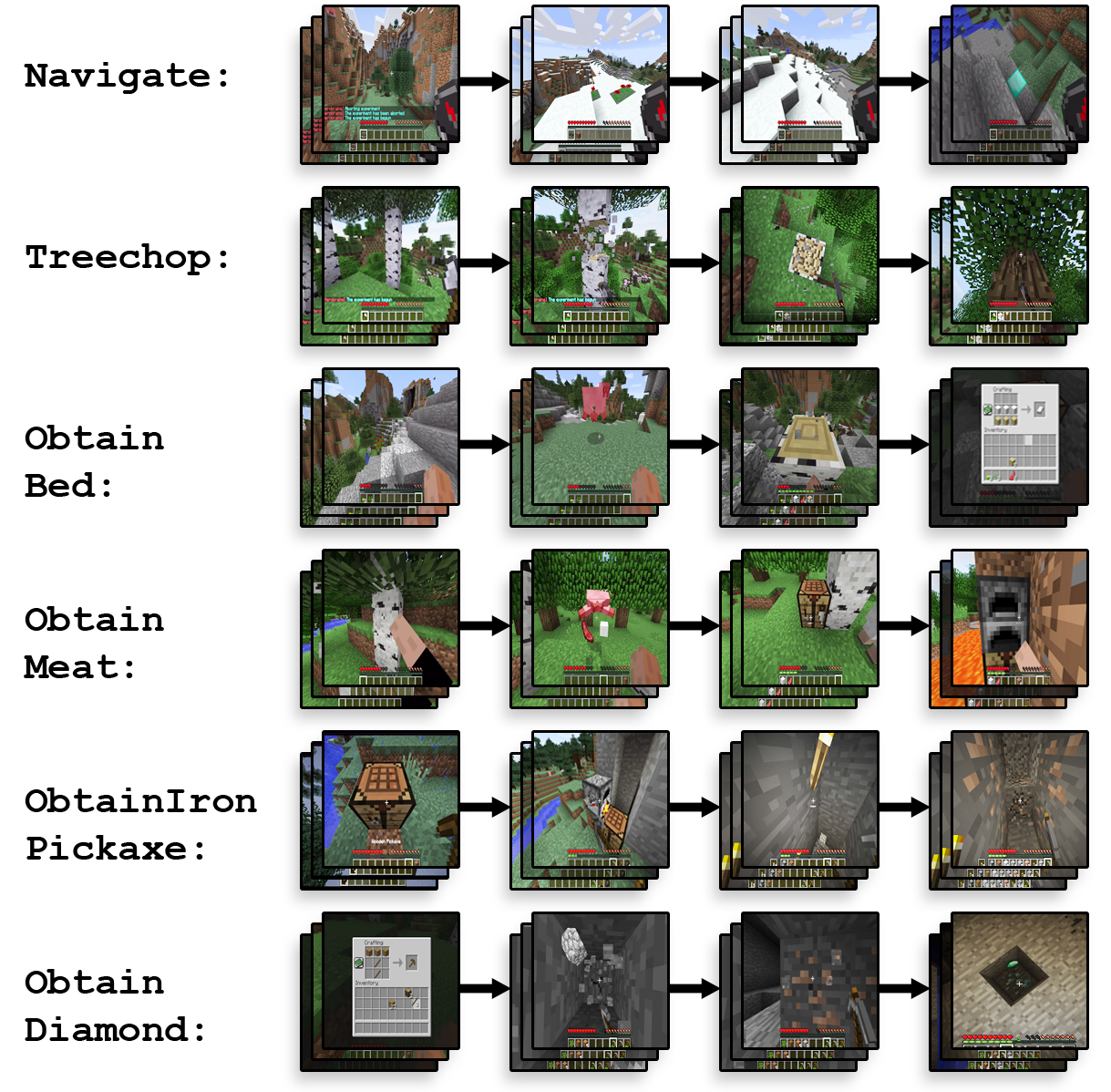}
        \caption{Images of various stages of the six stand-alone tasks (\texttt{Survial} gameplay not shown).}
        \vspace{-15pt}
    \end{center}
\end{figure}

\subsection{Dataset Details} \label{sec:data_details}

{
     \paragraph{Size.} The \minenet-v0 dataset consists of 500+ hours of recorded human demonstrations over six different tasks from the data collection platform. 
        The released data is comprised of four different versions of the dataset rendered with varied resolutions ($64\times64$ and $192\times256$) and textures (default Minecraft and simplified). 
        Each version individually totals to over 60 million state-action pairs with a size of 130 GB 
            and 734 GB for the low and medium resolution datasets respectively. 
}  
    
{
     \paragraph{Form.} Each trajectory is a contiguous set of state-action pairs sampled every Minecraft game tick (at 20 game ticks per second).
        Each state is comprised of an RGB video frame of the player's point-of-view and a comprehensive set of features from the game-state at that tick: player inventory, item collection events, distances to objectives, player attributes (health, level, achievements), and details about the current GUI the player has open. 
        The action recorded at each tick consists of: all of the keyboard presses on the client, the change in view pitch and yaw (caused by mouse movement), all player GUI click and interaction events, chat messages sent, and agglomorative actions such as item crafting.

    \paragraph{Additional Annotations.} Human trajectories are accompanied by a large set of automatically generated annotations. For all the stand-alone tasks, we record metrics which indicate the quality of the demonstration, such as timestamped rewards, number of no-ops, number of deaths, and total score. Additionally, the trajectory meta-data includes timestamped markers for hierarchical labelings; e.g. when a house-like structure is built or certain objectives such as chopping down a tree are met. 
}

{
    \paragraph{Packaging.} Each version of the dataset is packaged as a Zip archive with one folder per task family and one sub-folder per demonstration. In each trajectory folder, the states and actions are stored as an H.264 compressed MP4 video of the player's POV with a max bit rate of 18Mb/s and a JSON file containing all of the non-visual features of game-state as well as the player's actions corresponding to every frame of the video. Additionally, for specific \emph{task configurations} (simplifications of action and state space) we provide Numpy \texttt{.npz} files composed of state-action-reward tuples in vector form, promoting the accessibility of the dataset. The packaged data and accompanying documentation can be downloaded from \url{http://minerl.io}.
}

\input{sections/tasks.tex}
\begin{figure}
    \begin{center}
        \includegraphics[width=0.47\textwidth]{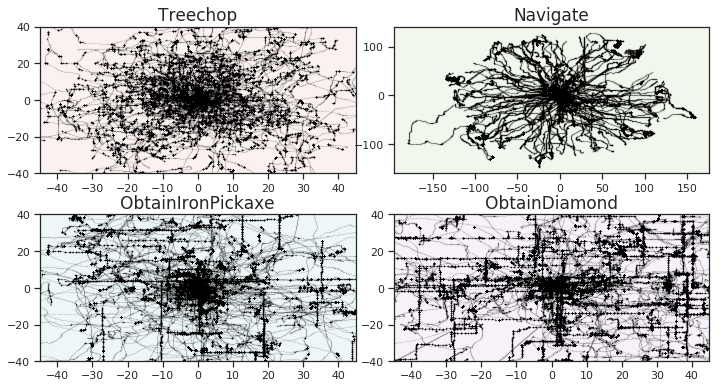}
    \end{center}
    \caption{  Plots of the XY positions of players in \texttt{Treechop}, \texttt{Navigate}, \texttt{ObtainIronPickaxe}, and \texttt{ObtainDiamond} overlaid so each player's individual, random initial location is $(0,0)$.
     }\label{fig:positions}
    \vspace{-15pt}
  
\end{figure}
\subsection{Analysis}

    \subsubsection{Human Performance}
    {
        A majority of the human demonstrations in the dataset fall squarely within expert level play.
            Figure \ref{fig:human_quality} shows the distribution over players of time required to complete each stand-alone task.
                The red region in each histogram denotes the range of times which correspond to play at an expert level, computed as the average time required for task completion by players with at least five years of Minecraft experience. 
                The large number of expert samples and rich labelings of demonstration 
                    performance enable application of many standard imitation learning techniques which assume optimality of the base policy. 
                In addition, the beginner and intermediate level trajectories 
                     allow for the further development of techniques which leverage imperfect demonstrations.
    }

    \subsubsection{Coverage} %
    { 
        \minenet-v0 has near complete coverage of Minecraft. 
            Within the \texttt{Survival} game mode, a large majority of the 371 subtasks for obtaining different items have been demonstrated by players hundreds to tens of thousands of times. 
                Further, some of these subtasks require hours to complete, requiring a long sequence of mining, building, exploring, and combatting enemies.  As a result of the large number of task-level annotations, the dataset can be used for large-scale option extraction and skill acquisition, enabling the extension of the work of \cite{shu2017hierarchical} and \cite{andreas2017modular}. Moreover, the rich label hierarchy of the \texttt{Obtain<Item>} tasks can be utilized in constructing metrics for the interpretability and quality of extracted options.

            In addition to item coverage, the \minenet{} data collection platform is structured to promote a broad representation of game conditions. The current dataset consists of a diverse set of demonstrations extracted from 1,002 unique player sessions. In the \texttt{Survival} game mode, the recorded trajectories collectively cover $24,393,057$ square meters of game content, where a square meter corresponds to one Minecraft block. For all other tasks, each demonstration occurs in a randomly initialized game world, so we collect a large number of unique, disparate trajectories for each task: 
                In Figure \ref{fig:positions}, we show the top-down position of players over the course of completing each task where the starting state is at $(0,0)$.
                Not only does each player act in a different game world, but each player also explores a large region during each task.
    }

    \subsubsection{Hierarchality} 
    { 
        As exemplified by the item graph shown in Figure \ref{fig:hierarchicality}, Minecraft is deeply hierarchical, and the \minenet~data collection platform is designed to capture these hierarchies both explicitly and implicitly. 
        As a primary example, the \texttt{Obtain<Item>} stand-alone tasks isolate difficult yet overlapping core paths in the item hierarchy.
        Due to the subtask labelings provided in \minenet-v0, we can inspect and quantify the extent to which these tasks overlap.

        A direct measure of hierachality emerges through \emph{item precedence frequency graphs}, graphs
            where nodes correspond to items obtained in a task and directed edges correspond to the number of times players obtained the source-node item immediately before the target-node item.
        
        These graphs provide a statistical view of the meta-policies of humans and the extent to which their subpolicies transfer between tasks. Figure \ref{fig:task_histogram} shows precedence frequency graphs constructed from \minenet{} trajectories on the \texttt{ObtainDiamond} , \texttt{ObtainCookedMeat} , and \texttt{ObtainIronPickaxe} tasks. 
                Inspection reveals that policies for obtaining a diamond consist of subpolicies which 
                    obtain wood, torches, and iron ore. All of these are also required for the \texttt{ObtainIronPickaxe} task, but only some of them are used within the \texttt{ObtainCookedMeat} task. 
                The effects of these overlapping subpolicies can be seen in Figure~\ref{fig:positions}: players move similarly in tasks with overlapping hierarchies (such as \texttt{ObtainIronPickaxe} and \texttt{ObtainDiamond} ) and move differently in tasks with less overlap.
                Moreover, these graphs paint a distributional picture of human meta-policies within a task: despite there being necessary graph traversal modes (e.g. wood $\to$ stone-pickaxe), depending on the situation, players adapt their strategies by acquiring items typically found later in the item precedence graph through longer paths when earlier items are unavailable. This, in turn, enables the use of \minenet{}-v0 in developing distributional hierarchical reinforcement learning methods.

                \begin{figure}
                    \begin{center}
                        \includegraphics[width=0.45\textwidth]{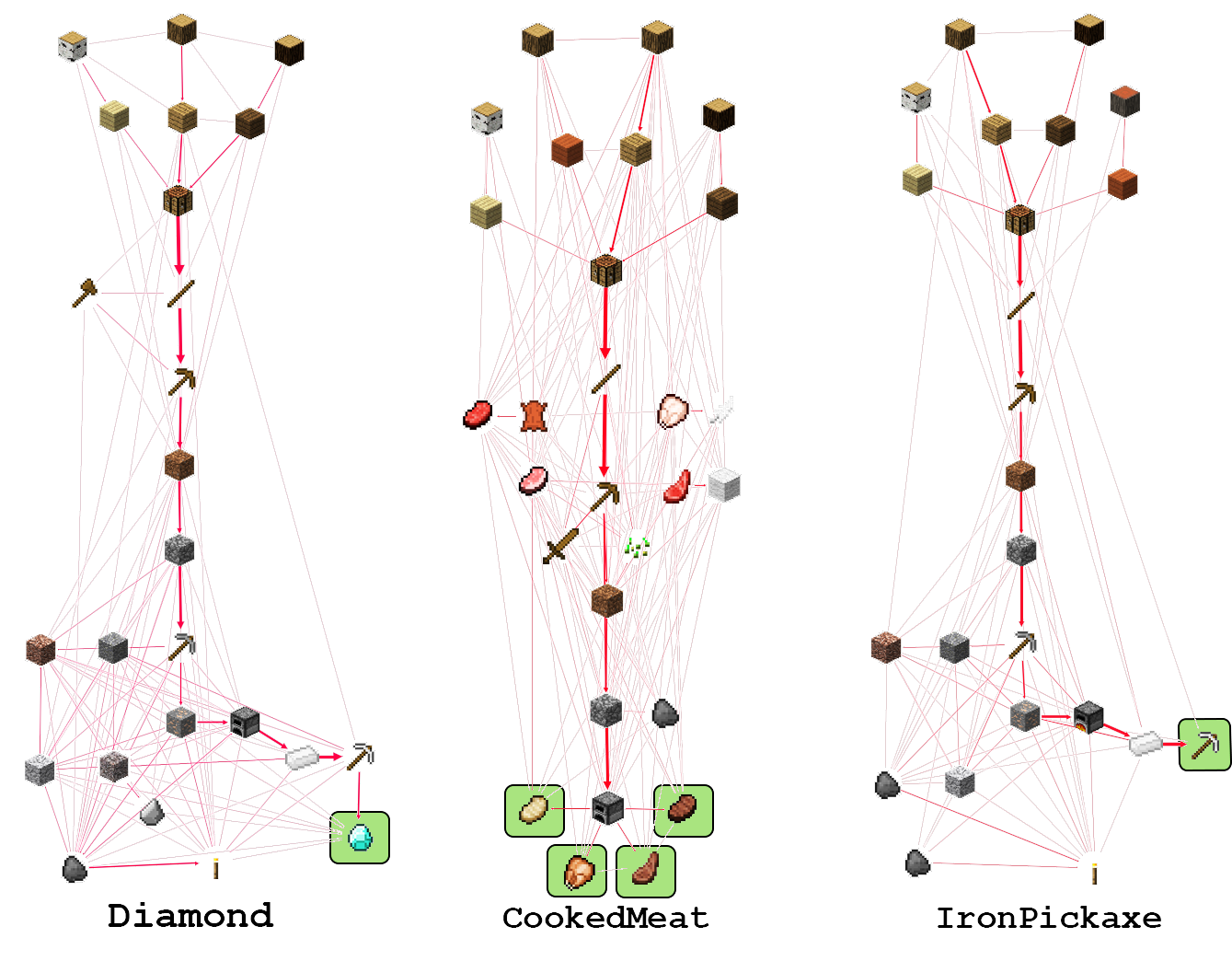} 
                        \caption{ Item precedence frequency graphs for \texttt{Obtain-} \texttt{Diamond} (left), \texttt{ObtainCookedMeat} (middle), and \texttt{ObtainIronPickaxe} (right). The thickness of each line indicates the number of times a player collected item $A$ then subsequently item $B$. } \label{fig:task_histogram}
                        \vspace{-18pt}
                    \end{center}
                \end{figure}

    }

 \section{Experiments} \label{sec:difficulty}
 \subsection{Experiment Configuration}

 To showcase the difficulty of Minecraft, we evaluate the performance of three reinforcement learning methods and one behavioral cloning method on the easiest of our tasks (\texttt{Treechop} and \texttt{Navigate} (Sparse)), as well as a simplified task with additional, shaped rewards, \texttt{Navigate} (Dense). Specifically, we evaluate (1) Dueling Double Deep Q-networks (DQN) \cite{mnih2015human}, an off-policy, Q-learning based method; (2) Pretrain DQN (PreDQN), DQN with additional pretraining steps and the replay buffer initialized with expert demonstrations from \minenet-v0; (3) Advantage Actor Critic (A2C) \cite{mnih2016asynchronous}, an on-policy, policy gradient method; and (4) Behavioral Cloning (BC), a method using standard classification techniques to learn a policy from demonstrations. To ensure reproducibility and an accurate evaluation of these methods, we build atop the OpenAI baseline implementations \cite{baselines}.

        Observations are converted to grey scale and resized to 64x64.
        Due to the thousands of action combinations in Minecraft and the limitations of the baseline algorithms, we simplify the action space to be 10 discrete actions.
        However, behavioral cloning does not have such limitations, and performs similarly without the action space simplifications.
        To use human demonstrations with Pretrained DQN and Behavioral Cloning, we approximate each action with one of our 10 action primitives.
        We train each reinforcement learning method for 1500 episode (approximately 12 million frames).
        To train Behavioral Cloning, we use expert trajectories from each respective task family and train until policy performance reaches its maximum. 
        
        \subsection{Evaluation and Discussion}
    We compare algorithms by the highest average reward obtained over a 100-episode window during training.
        We also report the performance of random policies and 50th percentile human performance.
        The results are summarized in Table~\ref{table:perf}.

    In all tasks, the learned agents perform significantly worse than human performance. \texttt{Treechop} exhibits the largest difference: humans achieve a score of 64, but reinforcement agents achieve scores of less than 4. This suggests that our tasks are quite difficult, especially given that the \texttt{Obtain<Item>} tasks build upon the \texttt{Treechop} task by requiring the completion of several additional subgoals ($\geq 3$).
        We hypothesize that a large source of difficulty comes from the environment's inherent long horizon credit assignment problems. For example, it is hard for agents to learn to navigate through water because it takes many transitions before the agent dies by drowning.

        In light of these difficulties, our data is useful in improving performance and sample efficiency: in all tasks, methods that leverage human data perform better. As seen in Figure~\ref{fig:dqn}, the expert demonstrations were able to achieve higher reward per episode and attain high performance using fewer samples.
        Expert demonstrations are particularly helpful in environments where random exploration is unlikely to yield any reward, like \texttt{Navigate} (Sparse).

    \begin{table}
            \small
                \begin{tabular}{lccc}
                    \toprule
                      & \texttt{Treechop} & \texttt{Navigate} (S) & \texttt{Navigate} (D) \\
                      \midrule
                     DQN & 3.73 $\pm$ 0.61 & 0.00 $\pm$ 0.00 & 55.59 $\pm$ 11.38 \\
                     A2C & 2.61 $\pm$ 0.50 & 0.00 $\pm$ 0.00 & -0.97 $\pm$ 3.23 \\
                     BC & \textbf{43.9} $\pm$ \textbf{31.46} & 4.23 $\pm$ 4.15 & 5.57 $\pm$ 6.00 \\
                    PreDQN & {4.16} $\pm$ {0.82} & \textbf{6.00} $\pm$ \textbf{4.65} & \textbf{94.96} $\pm$ \textbf{13.42} \\
                    \midrule
                    Human & 64.00 $\pm$ 0.00 & 100.00 $\pm$ 0.00 & 164.00 $\pm$ 0.00 \\
                    Random & 3.81 $\pm$ 0.57 & 1.00 $\pm$ 1.95 & -4.37 $\pm$ 5.10 \\
                    \bottomrule
                \end{tabular}
        \caption{
            Results in \texttt{Treechop}, \texttt{Navigate} (S)parse, and \texttt{Navigate} (D)ense, over the best 100 contiguous episodes. $\pm$ denotes standard deviation. 
            Note: humans achieve the maximum score for all tasks shown.
        }
        \label{table:perf}
        \vspace{-15pt}
    \end{table}

%% file: sections/tasks.tex
\subsection{Tasks}\label{sec:tasks}
    \begin{figure}
        \begin{center}
            \includegraphics[width=0.49\textwidth]{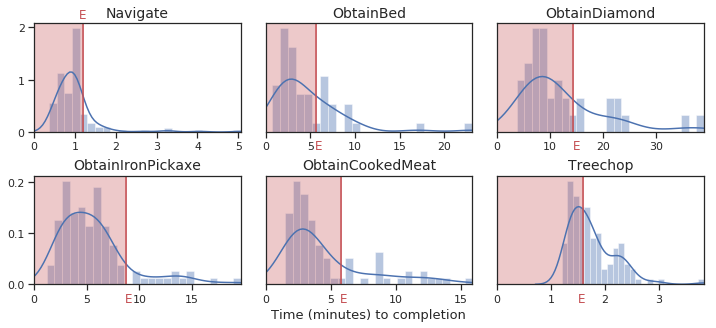}
        \caption{ Normalized histograms of the lengths of human demonstration on various \minenet~tasks. The red {\color{red} \tiny\sf{E}} denotes the upper threshold  for expert play on each task.}
        \label{fig:human_quality}
        \end{center}
        \vspace{-20pt}
    \end{figure}

\newcommand{\numBedVariants}{three}

The initial \minenet-v0 dataset consists of six stand-alone tasks chosen to represent difficult aspects of Minecraft that reflect challenges widely researched in the domain: hierarchality, long-term planning, and complex orienteering. Throughout all tasks, the agent has access to the same set of actions and observations as a human player, as outlined in Section \ref{sec:data_details}
All tasks have a time limit, which is part of the observation. Details for each task follow below.

\paragraph{Navigation.} In the \texttt{Navigate} task, the agent must move to a random goal location over procedurally generated, non-convex terrain with variable material type and geometry.
This is a subtask for many tasks throughout Minecraft. 
In addition to standard observations, the agent has access to a ``compass'' observation, which points to a set location, 64 blocks (meters) from the start location. 
The goal has a small random horizontal offset from this location and may be slightly below surface level, so the agent must find the final goal by searching based on visual features. 
There are two variants of the provided reward function: sparse (+1 upon reaching the goal, at which point the episode terminates), and dense (reward proportional to distance moved towards the goal). 

\paragraph{Tree Chopping.} The \texttt{Treechop} task replicates obtaining wood for producing further items. 
Wood is a key resource in Minecraft since it is a prerequisite for all tools (as seen by the placement of sticks in Figure~\ref{fig:hierarchicality} and Figure~\ref{fig:task_histogram}). 
The agent begins in a forest biome (near many trees) with an iron axe for cutting the trees. 
The agent is given +1 reward for obtaining each unit of wood, and the episode terminates once the agent obtains 64 units. 

\paragraph{Obtain Item.} We include four related tasks which require the agent to obtain an item further in the item hierarchy: \texttt{ObtainIronPickaxe}, \texttt{ObtainDiamond}, \texttt{ObtainCookedMeat}, and \texttt{ObtainBed}. 
The agent always begins in a random location without any items; this matches the starting conditions for human players in Minecraft. 
Different task variants correspond to a different, frequently used item: iron pickaxe, diamond, cooked meat (four variants, one per animal source), and bed (\numBedVariants{} variants, one per dye color needed). 
Iron pickaxes are tools required for obtaining key materials. 
Diamonds are central to high-level Minecraft play, and large portion of gameplay centers around their discovery. 
Cooked meat is used to replenish stamina, and a bed is required for sleeping.
Together, these items represent what a player would need to obtain to survive and access further areas of the game. The agent is given $+1$ reward for obtaining the required item, at which point the episode terminates.

\paragraph{Survival.} 
In addition to data on specific, designed tasks, we provide data in \texttt{Survival}, the standard open-ended game mode used by most players. 
Starting from a random location without any items, players formulate their own high-level goals and obtain items to complete these goals.
Data from this task can be used for learning the intricate reward functions followed by humans in open play and the corresponding policies. 
This data can also be used to train agents attempting to complete the other, structured tasks, or further for extracting policy sketches as in \cite{andreas2017modular}.

%% file: sections/related_works.tex

%% file: sections/conclusion.tex

\vspace{-7pt}
\section{Related Work}

    A number of domains have been previously solved through imitation learning and a dataset of human demonstrations. These include the Atari domain using the Atari Grand Challenge dataset~\cite{kurin2017atari} and the Super Tux Kart domain using an on-demand dataset~\cite{ross2011reduction}. Unlike Minecraft, these are simple domains: they have shallow dependency hierarchies and are not open-world. Due to the small action- and state-spaces, these domains have been solved using imitation learning using relatively few samples (9.7 million frames across five games in~\cite{kurin2017atari} and 20 thousand frames in~\cite{ross2011reduction}). In contrast, we present 60 million automatically annotated state-action pairs and do not achieve human performance. 

    Existing datasets for challenging, unsolved domains are primarily for real-world tasks where a lack of simulators limits the pace of development. The KITTI dataset \cite{geiger2013vision}, for example, is a dataset of 3 hours of 3D information on real-world traffic. Similarly, Dex-Net \cite{mahler2019learning} is a dataset of five million grasps with corresponding 3D pointclouds for robotic manipulation. Unlike these datasets, \minenet{} is directly compatible with a simulator, Malmo, thereby allowing training in the same domain as the data was gathered and comparison to methods not based on imitation learning. Additionally, the scale of \minenet{} is larger relative to the domain difficulty than the KITTI and Dex-Net datasets.

    The only complex, unsolved domain with an existing simulator and large-scale dataset is StarCraft~II. However, StarCraft~II is not open-world so cannot be used to evaluate methods designed for embodied tasks in 3D environments. The largest dataset is currently StarData \cite{lin2017stardata}. Unlike \minenet{}, it consists of unlabeled, extracted trajectories of standard gameplay. In contrast, \minenet{} includes a growing number of related tasks which represent different components of the overall Minecraft task hierarchy. In addition, \minenet{} consists of rich automatically generated annotations including subtask completion, player skill-level, and an API to extend these labels. Together, these properties allow the use and evaluation of methods which exploit hierarchical structures.

    \begin{figure}
        \begin{center}
            \includegraphics[width=0.48\textwidth]{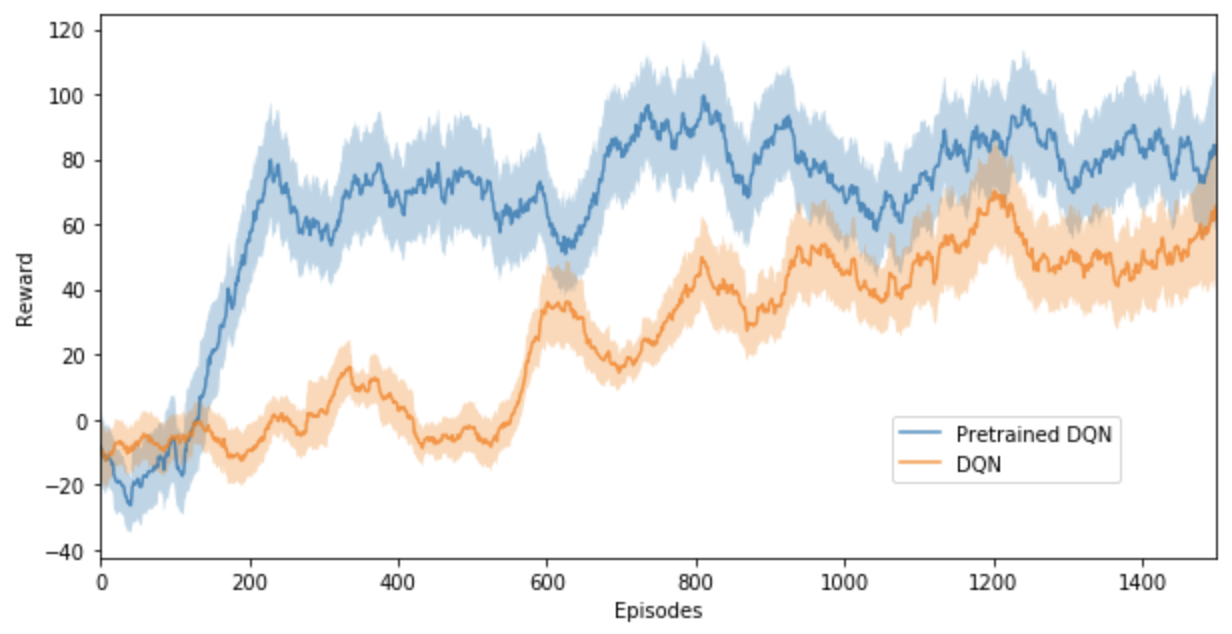} 
            \caption{ Performance graphs over time with DQN and pretrained DQN on \texttt{Navigate} (Dense).} \label{fig:dqn}
            \vspace{-20pt}
        \end{center}
    \end{figure}

\vspace{-7pt}
\section{Conclusion and Future Work}

    \minenet-v0 currently features 60 million state-action pairs of procedurally annotated human demonstrations in an open-world, simulator-paired environment. 
        It currently contains data for six tasks, none of which can be fully solved with standard deep reinforcement learning methods. 
    Our platform allows for the ongoing collection of demonstrations for both existing and new tasks.
        Thus, we host \minenet-v0 at a community accessible website, \url{http://minerl.io}, and will gather feedback on adding new annotations and tasks to \minenet{}.
        As we expand \minenet{}, we expect it to be increasingly useful for a range of methods including inverse reinforcement learning, hierarchical learning, and life-long learning.
    We hope \minenet{} will become a central resource for sequential decision-making research, bolstering many branches of AI toward the common goal of developing methods capable of solving a wider range of real-world environments.

\section*{Acknowledgements}

We would like to thank Greg Yang, Devendra Chaplot, Lucy Cheung,  Stephanie Milani, Miranda Chen, Yiwen Yuan, Cheri Guss, Steve Shalongo, Jim Guss, Sauce, and Bridget Hickey for their insightful conversations and support.

%% file: ijcai19.bbl
\begin{thebibliography}{}

\bibitem[\protect\citeauthoryear{Andreas \bgroup \em et al.\egroup
  }{2017}]{andreas2017modular}
Jacob Andreas, Dan Klein, and Sergey Levine.
\newblock Modular multitask reinforcement learning with policy sketches.
\newblock In {\em Proceedings of the 34th ICML-Volume 70}, pages 166--175.
  JMLR. org, 2017.

\bibitem[\protect\citeauthoryear{Andrychowicz \bgroup \em et al.\egroup
  }{2018}]{andrychowicz2018learning}
Marcin Andrychowicz, Bowen Baker, Maciek Chociej, Rafal Jozefowicz, Bob McGrew,
  Jakub Pachocki, Arthur Petron, Matthias Plappert, Glenn Powell, Alex Ray,
  et~al.
\newblock Learning dexterous in-hand manipulation.
\newblock {\em arXiv preprint arXiv:1808.00177}, 2018.

\bibitem[\protect\citeauthoryear{Bellemare \bgroup \em et al.\egroup
  }{2013}]{bellemare2013arcade}
Marc~G Bellemare, Yavar Naddaf, Joel Veness, and Michael Bowling.
\newblock The arcade learning environment: An evaluation platform for general
  agents.
\newblock {\em JAIR}, 47:253--279, 2013.

\bibitem[\protect\citeauthoryear{DeepMind}{2018}]{deepmind}
DeepMind.
\newblock Alphastar: Mastering the real-time strategy game starcraft ii, 2018.

\bibitem[\protect\citeauthoryear{Deng \bgroup \em et al.\egroup
  }{2009}]{deng2009imagenet}
Jia Deng, Wei Dong, Richard Socher, Li-Jia Li, Kai Li, and Li~Fei-Fei.
\newblock Imagenet: A large-scale hierarchical image database.
\newblock 2009.

\bibitem[\protect\citeauthoryear{Dhariwal \bgroup \em et al.\egroup
  }{2017}]{baselines}
Prafulla Dhariwal, Christopher Hesse, Oleg Klimov, Alex Nichol, Matthias
  Plappert, Alec Radford, John Schulman, Szymon Sidor, Yuhuai Wu, and Peter
  Zhokhov.
\newblock Openai baselines, 2017.

\bibitem[\protect\citeauthoryear{Geiger \bgroup \em et al.\egroup
  }{2013}]{geiger2013vision}
Andreas Geiger, Philip Lenz, Christoph Stiller, and Raquel Urtasun.
\newblock Vision meets robotics: The kitti dataset.
\newblock {\em IJRR}, 32(11):1231--1237, 2013.

\bibitem[\protect\citeauthoryear{Godfrey \bgroup \em et al.\egroup
  }{1992}]{godfrey1992switchboard}
John~J Godfrey, Edward~C Holliman, and Jane McDaniel.
\newblock Switchboard: Telephone speech corpus for research and development.
\newblock In {\em Acoustics, Speech, and Signal Processing, 1992. ICASSP-92.,
  1992 IEEE International Conference on}, volume~1, pages 517--520. IEEE, 1992.

\bibitem[\protect\citeauthoryear{Hessel \bgroup \em et al.\egroup
  }{2018}]{hessel2018rainbow}
Matteo Hessel, Joseph Modayil, Hado Van~Hasselt, Tom Schaul, Georg Ostrovski,
  Will Dabney, Dan Horgan, Bilal Piot, Mohammad Azar, and David Silver.
\newblock Rainbow: Combining improvements in deep reinforcement learning.
\newblock In {\em Thirty-Second AAAI Conference on Artificial Intelligence},
  2018.

\bibitem[\protect\citeauthoryear{Johnson \bgroup \em et al.\egroup
  }{2016}]{johnson2016malmo}
Matthew Johnson, Katja Hofmann, Tim Hutton, and David Bignell.
\newblock The malmo platform for artificial intelligence experimentation.
\newblock In {\em IJCAI}, pages 4246--4247, 2016.

\bibitem[\protect\citeauthoryear{Kurin \bgroup \em et al.\egroup
  }{2017}]{kurin2017atari}
Vitaly Kurin, Sebastian Nowozin, Katja Hofmann, Lucas Beyer, and Bastian Leibe.
\newblock The atari grand challenge dataset.
\newblock {\em arXiv preprint arXiv:1705.10998}, 2017.

\bibitem[\protect\citeauthoryear{Levine \bgroup \em et al.\egroup
  }{2018}]{levine2018learning}
Sergey Levine, Peter Pastor, Alex Krizhevsky, Julian Ibarz, and Deirdre
  Quillen.
\newblock Learning hand-eye coordination for robotic grasping with deep
  learning and large-scale data collection.
\newblock {\em IJRR}, 37(4-5):421--436, 2018.

\bibitem[\protect\citeauthoryear{Lin \bgroup \em et al.\egroup
  }{2017}]{lin2017stardata}
Zeming Lin, Jonas Gehring, Vasil Khalidov, and Gabriel Synnaeve.
\newblock Stardata: A starcraft ai research dataset.
\newblock In {\em Thirteenth AIDE Conference}, 2017.

\bibitem[\protect\citeauthoryear{Liu \bgroup \em et al.\egroup
  }{2017}]{liu2017interactive}
Jerry Liu, Fisher Yu, and Thomas Funkhouser.
\newblock Interactive 3d modeling with a generative adversarial network.
\newblock In {\em 2017 IC3DV}, pages 126--134. IEEE, 2017.

\bibitem[\protect\citeauthoryear{Mahler \bgroup \em et al.\egroup
  }{2019}]{mahler2019learning}
Jeffrey Mahler, Matthew Matl, Vishal Satish, Michael Danielczuk, Bill DeRose,
  Stephen McKinley, and Ken Goldberg.
\newblock Learning ambidextrous robot grasping policies.
\newblock {\em Science Robotics}, 4(26):eaau4984, 2019.

\bibitem[\protect\citeauthoryear{Mnih \bgroup \em et al.\egroup
  }{2015}]{mnih2015human}
Volodymyr Mnih, Koray Kavukcuoglu, David Silver, Andrei~A Rusu, Joel Veness,
  Marc~G Bellemare, Alex Graves, Martin Riedmiller, Andreas~K Fidjeland, Georg
  Ostrovski, et~al.
\newblock Human-level control through deep reinforcement learning.
\newblock {\em Nature}, 518(7540):529, 2015.

\bibitem[\protect\citeauthoryear{Mnih \bgroup \em et al.\egroup
  }{2016}]{mnih2016asynchronous}
Volodymyr Mnih, Adria~Puigdomenech Badia, Mehdi Mirza, Alex Graves, Timothy
  Lillicrap, Tim Harley, David Silver, and Koray Kavukcuoglu.
\newblock Asynchronous methods for deep reinforcement learning.
\newblock In {\em ICML}, pages 1928--1937, 2016.

\bibitem[\protect\citeauthoryear{Oh \bgroup \em et al.\egroup
  }{2016}]{oh2016control}
Junhyuk Oh, Valliappa Chockalingam, Satinder Singh, and Honglak Lee.
\newblock Control of memory, active perception, and action in minecraft.
\newblock {\em arXiv preprint arXiv:1605.09128}, 2016.

\bibitem[\protect\citeauthoryear{OpenAI}{2018}]{openai_2018}
OpenAI.
\newblock Openai five, Sep 2018.

\bibitem[\protect\citeauthoryear{Ross \bgroup \em et al.\egroup
  }{2011}]{ross2011reduction}
St{\'e}phane Ross, Geoffrey Gordon, and Drew Bagnell.
\newblock A reduction of imitation learning and structured prediction to
  no-regret online learning.
\newblock In {\em Proceedings of the 14th ICIAS}, pages 627--635, 2011.

\bibitem[\protect\citeauthoryear{Salge \bgroup \em et al.\egroup
  }{2014}]{salge2014changing}
Christoph Salge, Cornelius Glackin, and Daniel Polani.
\newblock Changing the environment based on empowerment as intrinsic
  motivation.
\newblock {\em Entropy}, 16(5):2789--2819, 2014.

\bibitem[\protect\citeauthoryear{Shu \bgroup \em et al.\egroup
  }{2017}]{shu2017hierarchical}
Tianmin Shu, Caiming Xiong, and Richard Socher.
\newblock Hierarchical and interpretable skill acquisition in multi-task
  reinforcement learning.
\newblock {\em arXiv preprint arXiv:1712.07294}, 2017.

\bibitem[\protect\citeauthoryear{Silver \bgroup \em et al.\egroup
  }{2017}]{silver2017mastering}
David Silver, Julian Schrittwieser, Karen Simonyan, Ioannis Antonoglou, Aja
  Huang, Arthur Guez, Thomas Hubert, Lucas Baker, Matthew Lai, Adrian Bolton,
  et~al.
\newblock Mastering the game of go without human knowledge.
\newblock {\em Nature}, 550(7676):354, 2017.

\bibitem[\protect\citeauthoryear{Tessler \bgroup \em et al.\egroup
  }{2017}]{tessler2017deep}
Chen Tessler, Shahar Givony, Tom Zahavy, Daniel~J Mankowitz, and Shie Mannor.
\newblock A deep hierarchical approach to lifelong learning in minecraft.
\newblock In {\em Thirty-First AAAI}, 2017.

\bibitem[\protect\citeauthoryear{Tobin \bgroup \em et al.\egroup
  }{2017}]{tobin2017domain}
Josh Tobin, Rachel Fong, Alex Ray, Jonas Schneider, Wojciech Zaremba, and
  Pieter Abbeel.
\newblock Domain randomization for transferring deep neural networks from
  simulation to the real world.
\newblock In {\em Intelligent Robots and Systems (IROS), 2017 IEEE/RSJ
  International Conference on}, pages 23--30. IEEE, 2017.

\bibitem[\protect\citeauthoryear{Wang \bgroup \em et al.\egroup
  }{2018}]{wang2018pix2pixHD}
Ting-Chun Wang, Ming-Yu Liu, Jun-Yan Zhu, Andrew Tao, Jan Kautz, and Bryan
  Catanzaro.
\newblock High-resolution image synthesis and semantic manipulation with
  conditional gans.
\newblock In {\em Proceedings of the IEEE Conference on Computer Vision and
  Pattern Recognition}, pages 8798--8807, 2018.

\end{thebibliography}
